\documentclass[runningheads]{llncs}
\usepackage{graphicx}

\usepackage{cite}
\usepackage{amsmath,amssymb,amsfonts}
\usepackage{algorithmic}
\usepackage{graphicx}
\usepackage{textcomp}
\usepackage{xcolor}
\usepackage{pifont}

\usepackage{todonotes}
\usepackage{subcaption}
\usepackage{accents} 
\usepackage{multirow}
\usepackage{url}
\usepackage{svg}

\usepackage{textcomp} 
\usepackage{pifont} 

\newcommand{\equalcontribution}{\textsuperscript{\ding{70}}}

\DeclareMathOperator*{\argmin}{arg\,min}

\begin{document}
\title{An Efficient Multi Quantile Regression Network with Ad Hoc Prevention of Quantile Crossing\thanks{This research has been funded by the Federal Ministry for Economic Affairs and Climate Action (BMWK) within the project "KI-basierte Topologieoptimierung elektrischer Maschinen (KITE)" (19I21034C).}}
\titlerunning{Efficient Multi Quantile Regression Network}
%
\author{Jens Decke\equalcontribution\orcidID{0000-0002-7893-1564} \and
Arne Jenß\equalcontribution\orcidID{0009-0007-7609-3783} \and \\
Bernhard Sick\orcidID{0000-0001-9467-656X} \and
Christian Gruhl\orcidID{0000-0001-9838-3676}}
\authorrunning{J. Decke et al.}
%
\institute{Intelligent Embedded Systems, University of Kassel, 34121 Kassel, Germany \\
\email{\{jdecke, arne.jenss, bsick, cgruhl\}@uni-kassel.de}\\
\url{https://www.uni-kassel.de/eecs/ies/}\\
\textsuperscript{\ding{70}}Equally contributed.}
\maketitle
\begin{abstract}
This article presents the Sorting Composite Quantile Regression Neural Network (SCQRNN), an advanced quantile regression model designed to prevent quantile crossing and enhance computational efficiency. Integrating ad hoc sorting in training, the SCQRNN ensures non-intersecting quantiles, boosting model reliability and interpretability. We demonstrate that the SCQRNN not only prevents quantile crossing and reduces computational complexity but also achieves faster convergence than traditional models. This advancement meets the requirements of high-performance computing for sustainable, accurate computation. In organic computing, the SCQRNN enhances self-aware systems with predictive uncertainties, enriching applications across finance, meteorology, climate science, and engineering.

\end{abstract}

\section{Introduction}

    In the field of organic computing, quantile regression aligns with the core principles, including self-awareness and self-adap\-tation. This method integrates well with the self-organizing nature of organic systems, effectively responding to scenarios with varying degrees of uncertainty. In this context, quantile regression is an example of organic computing's goal for efficient computing and a base concept for self-aware systems, where we model the environment with different degrees of uncertainty~\cite{muller2011organic,gruhl2015A}.

    Quantile regression has become an indispensable tool in statistical analysis, allowing for a more comprehensive understanding of the conditional distribution of a response variable. Unlike mean regression, which offers a singular view of central tendency, quantile regression provides a richer, more nuanced depiction of potential outcomes by estimating conditional quantile functions. This technique is particularly valuable in fields where understanding the variability of predictions is as crucial as the predictions themselves, such as in economics, finance, meteorology and engineering.

    However, a persistent challenge in quantile regression is the phenomenon of quantile crossing, where estimated quantiles may intersect, leading to a violation of the basic principle that higher quantiles must be greater than (or equal to) lower quantiles. This issue not only disrupts the interpretability of the regression model but also undermines the reliability of the inference drawn from it.


    Previous attempts to address quantile crossing often come with a significant computational cost, or rely heavily on post-processing. These methods can be particularly burdensome in scenarios involving large datasets or the need for real-time analysis. Moreover, the complexity of these solutions can pose barriers to their practical implementation in various applied settings.

    In the realm of high-performance computing (HPC), the intersection of computational efficiency and sustainable computing has become increasingly critical. As we delve deeper into the complexities of machine learning and statistical analysis, the environmental implications of these computationally intensive processes, particularly regarding energy consumption and associated greenhouse gas emissions, cannot be ignored. This is particularly relevant in the field of neural network quantile regression, where the need for processing power has traditionally led to significant energy use, raising concerns over ecological impact.

    In this study, we present an innovative approach designed to address the issue of quantile crossing in quantile regression models. Our method: Sorting Composite Quantile Regression Neural Network (SCQRNN), is centered around a novel algorithmic solution that seamlessly integrates with the quantile regression framework. Its primary strength lies in its computational efficiency, which significantly reduces both time and resources required, maintaining accuracy and robustness. The major contributions of our work are outlined as follows:
    \begin{itemize}
        \item Development of a more efficient model for non-crossing quantile regression~\footnote{\url{https://gitlab.uni-kassel.de/uk045707/scqrnn}}.
        \item Theoretical complexity analysis of our proposed method.
        \item Comparative analysis of our approach against state-of-the-art models using nine datasets, evaluating the root mean squared error and overall reliability.
        \item Investigation of convergence speed compared to a reasonable baseline evaluated on a real-world problem.
    \end{itemize}


\section{Related Work}\label{related_work}
    The concept of quantile estimation through regression traces back to the pioneering work of Koenker and Bassett in 1978~\cite{koenker1978regression}. For a given $\tau \in (0,1)$, consider  $y^\tau$ as the $\tau$th quantile  of a random sample $\{y_i:i\in 1, ..., N\}$ on a random Variable $Y$. Koenker and Bassett~\cite{koenker1978regression} use the fact, that $y^\tau$ can be described as the solution of the following minimization problem:

    \begin{equation}\label{min_eq}
        y^\tau=\argmin_{\hat{y}^\tau\in \mathbb{R}}[\sum_{i\in\{i:y_i\geq\hat{y}^\tau\}}\tau |y_i-\hat{y}^\tau|+\sum_{i\in\{i:y_i<\hat{y}^\tau\}}(1-\tau) |y_i-\hat{y}^\tau|]
    \end{equation}

    While they only construct a simple linear model for their regression, this is the exact same concept, which is used today, to design loss functions for quantile regression neural networks. 

    \subsection{Quantile Regression Neural Network}
        In 2011 Cannon \cite{cannon2011quantile} introduced the use of the checker function 
        
        \begin{equation}\label{checker_eq}
            \rho_\tau(u)=
            \begin{cases}
                \tau u & \text{if $u\geq0$} \\
                (\tau-1)u & \text{if $u<0$}
            \end{cases}
        \end{equation}
        
        to formulate the loss function 
        
        \begin{equation}\label{error_eq}
            EQ_\tau=\frac{1}{N} \sum\limits_{i=1}^{N} \rho_\tau(y_i-\hat{y}^\tau_i)
        \end{equation}

        for the minimization problem in (\ref{min_eq}), a method also known as the pinball loss.
        Due to the non-differentiability of the checker function (\ref{checker_eq}) at $u = 0$, a modified version is used to train the Quantile Regression Neural Network(QRNN). 
         To achieve this, the Huber norm, proposed by Huber\cite{huber1964robust} in 1964, is used
        to create a modified checker function, that is differentiable.

        The QRNN model developed in \cite{cannon2011quantile}  operates as a multilayer perceptron with a single output neuron, that is capable to predict a single specific quantile function. Consequently, to predict multiple quantiles -for instance to predict confidence intervals- seperate models must be trained for each desired quantile. This approach is not only inefficient but also does not prevent the potential crossing of the predicted quantile functions.

    \subsection{Composite Quantile Regression Neural Network}\label{CQRNN}
        A method to predict multiple quantiles with a single model, is introduced by Xu et al. in 2017 \cite{xu2017composite}. This model also resembles a multilayer perceptron, but with  $T$ output neurons for each of the $\mathbf{\tau}=(\tau_1,...,\tau_T)$ wanted quantiles for prediction. As a result, this necessitates another error function, essentially an average of the loss function (\ref{error_eq}) evaluated individually for each $\tau_k$:
        
        \begin{equation}\label{error3_eq}
            ECQ_\mathbf{\tau}=\frac{1}{T} \sum\limits_{k=1}^{T} EQ_{\tau_k} =\frac{1}{TN} \sum\limits_{k=1}^{T} \sum\limits_{i=1}^{N} \rho_{\tau_k}(y_i-\hat{y}^{\tau_k}_i)
        \end{equation}

        For $T=1$ the Composite Quantile Regression Neural Network (CQRNN) is identical to the QRNN.
        Unfortunately it also might suffer from quantile crossing. In section \ref{methodology} we will use the CQRNN as a basis for the SCQRNN.
    
    \subsection{Monotone Composite Quantile Regression Neural Network}\label{MCQRNN}
        Cannon, in 2017~\cite{Cannon2017non}, offered a solution to the issue of quantile crossing by incorporating monotone constraints within a neural network, a concept initially outlined by Zhang~\cite{zhang1999feedforward} on feedforward networks.
         These monotone constraints make it possible to guarantee a monotone relationship between certain features of the input vector $x\in\mathbb{R}^M$ and the output variable $y\in\mathbb{R}$ of a neural network. In detail this is achieved by manipulating the weights of the input layer by feeding them into an exponential function.
        
        Assume without loss of generality that the first $m$ features of $x$ are those that must be monotone in the output. Then the output of the first layer $z_1$ with weight matrix $W^{(1)}$, bias $b^{(1)}$ and activation function $f$ is denoted as follows:

        \begin{equation}\label{monotone_input_eq}
            z^{(1)} = f(\sum\limits_{i=1}^m \exp{(W_i^{(1)})} x_i + \sum\limits_{j=m+1}^M W_{j}^{(1)} x_j + b^{(1)})
        \end{equation}

        To preserve the monotonicity established in the first layer, the exponential function is applied to the whole weight matrices in the following layers:

        \begin{equation}\label{monotone_layer_eq}
            z^{(k)} = f(\exp{(W^{(k)})}z^{(k-1)} + b^{(k)})
        \end{equation}

        It's important to clarify that the $\exp$-function mentioned in both Equations~(\ref{monotone_input_eq}) and (\ref{monotone_layer_eq}) refers to the exponential function that is applied element wise, rather than the exponential of a matrix.

         When the predictions $\hat{y}^\mathbf{\tau}$ for the set of quantiles $\mathbf{\tau}=(\tau_1,...,\tau_T)$ exhibit quantile crossing, it implies a lack of monotonicity with respect to $\mathbf{\tau}$. To address this, Cannon`s Monotone Composite Quantile Regression Neural Network (MCQRNN) introduces monotone constraints to ensure the predictions are monotone across all quantiles~\cite{Cannon2017non}. Consequently, $\mathbf{\tau}$ is integrated into the design matrix, which is processed by a neural network with monotone constraints in $\mathbf{\tau}$.

        Assume the original data is given by the matrix $X\in\mathbb{R}^{M\times N}$ and the target vector $y\in\mathbb{R}^N$. Then the design matrix and target vector of the MCQRNN is expressed as follows:

        \begin{equation} \label{design_matrix}
            \tilde{X}=
            \begin{bmatrix}
                \tau_1 & \cdots & \tau_1 &  \cdots & \tau_T & \cdots & \tau_T \\
                x_{11} & \cdots & x_{N1} &  \cdots & x_{11} & \cdots & x_{N1} \\
                \vdots & \ddots & \vdots &  \ddots & \vdots & \ddots & \vdots \\
                x_{1M} & \cdots & x_{NM} &  \cdots & x_{1M} & \cdots & x_{NM}
            \end{bmatrix}
            , \tilde{y}=
            \begin{bmatrix}
                y_1\\
                \vdots\\
                y_N\\
                \vdots\\
                y_1\\
                \vdots\\
                y_N
            \end{bmatrix}
        \end{equation}

        The matrix $\tilde{X}\in\mathbb{R}^{M+1\times TN}$ results from concatenating the $X$ matrix $T$ times  and adding an additional feature for the $\mathbf{\tau}$ values. The target vector $\tilde{y}$ is simply the original $y$ repeated $T$ times. This expanded matrices are then used for supervised training as described in Equation~\eqref{monotone_input_eq} and Equation~\eqref{monotone_layer_eq} with at least the feature $\mathbf{\tau}$ as monotone (note, that monotone constraints can still be added for additional features).
        The error function used is essentially the same as for the QRNN Equation~\eqref{error_eq} with the critical difference, that during learning the $\mathbf{\tau}$ values must be passed alongside the predictions. This adjustment allows the loss to be tailored for different $\mathbf{\tau}$ values.

    \subsection{ Differentiable Sorting}\label{diff_sort}

        Fakoor~et~al.~\cite{fakoor2023flexible} described sorting as a possible post hoc adjustment for multi-quantile estimation to achieve noncrossing quantiles. They also demonstrated, that applying post hoc sorting enhances the pinball loss detailed in Equation~\eqref{error_eq}:

        \begin{proposition}\label{prop:pinball}
            Let $\hat{y}^{\mathbf{\tau}}=(\hat{y}^{\tau_1},...,\hat{y}^{\tau_T})$ be an estimate of the conditional quantile function at a point $x$ for $\mathbf{\tau}=(\tau_1,...,\tau_T)$. Let $\check{y}^{\mathbf{\tau}}=\mathcal{S}(\hat{y}^{\mathbf{\tau}})$ with $\mathcal{S}$ being the sorting operator. Then the following holds for for any $y\in\mathbb{R}$:
            \begin{equation*}
                \sum\limits_{k=1}^{T} \rho_{\tau_k}(y-\check{y}^{\tau_k})\leq\sum\limits_{k=1}^{T} \rho_{\tau_k}(y-\hat{y}^{\tau_k})
            \end{equation*}
            Moreover, if sorting is nontrivial: $\check{y}^{\mathbf{\tau}}\neq\hat{y}^{\mathbf{\tau}}$ the inequality is strict.
        \end{proposition}

        A proof for this proposition is also provided in \cite{fakoor2023flexible}.

        The SCQRNN model introduced in section \ref{methodology} utilizes the differentiable sorting algorithm proposed by Blondel~et~al.~\cite{blondel2020fast}. Their method achieved a $\mathcal{O}(n\log{n})$ computation complexity and a $\mathcal{O}(n)$ differentiation complexity, which makes it suitable for application during optimization.


\section{Methodology}\label{methodology}
    
    We modify the CQRNN  approach by Xu et al.\cite{xu2017composite} further, to include ad hoc sorting during the training of the model.

    \subsection{Sorted Composite Quantile Regression Neural Network}\label{SCQRNN}
        Let $\mathbf{\tau}=(\tau_1,...,\tau_T)$ be our quantiles, $x\in\mathbb{R}^M$ the input vector and $y\in\mathbb{R}$  the output variable. For our model design, we additionally need an actitivation function $f$ and the integer vector $\kappa=(\kappa_1,...,\kappa_K)\in\mathbb{N}^K_+$, which describes the shapes of our hidden layers. This yields us the $K+1$ weight matrices

        \begin{equation}
            W^{(k)}\in
            \begin{cases}
                \mathbb{R}^{\kappa_1\times M} & \text{if $k=0$} \\
                \mathbb{R}^{\kappa_{k+1}\times\kappa_{k}} & \text{if $0<k<K$} \\
                \mathbb{R}^{T \times\kappa_K} & \text{if $k=K$}
            \end{cases}
        \end{equation}

        and bias vectors $b^{(k)}\in\mathbb{R}^{\kappa_k}$, $b^{(K)}\in\mathbb{R}^T$.
        We then calculate 
        \begin{equation}
            z^{(k)}=f(W^{(k-1)}~z^{(k-1)}~+~b^{(k)})
        \end{equation}
        iterative with $z^{(0)}=x$. The output of our forward pass is then given by 
        \begin{equation}
            \hat{y}^\mathbf{\tau}~=~\mathcal{S}(z^{(K+1)})
        \end{equation}
        where $\mathcal{S}$ denotes the sorting operation.

        Since we use the implementation from Blondel~et~al.~\cite{blondel2020fast} for sorting, we know that $\mathcal{S}$ is differentiable. Therefore, $\mathcal{S}$ can be regarded an additional layer without trainable weights in the optimization. For the optimization itself we use the Adam algorithm by Kingma~and~Ba~\cite{kingma2017adam} to optimize the loss function given in Equation \eqref{error3_eq}. 

        \begin{figure}[b!]
            \centering
            \includegraphics[width=0.6\linewidth]{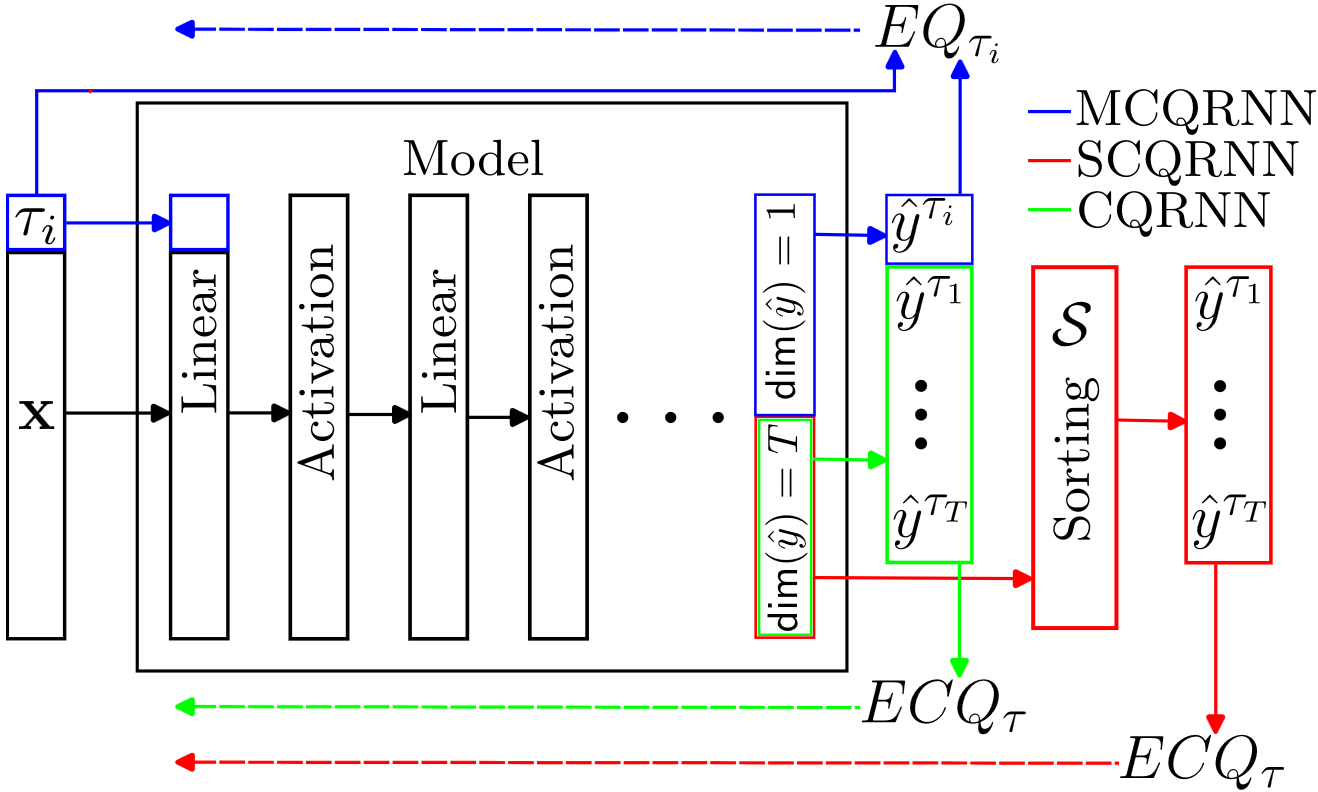}
            \caption{Illustration of the MCQRNN, SCQRNN and CQRNN}
            \label{fig:model_illustration}
         \end{figure}

        Figure \ref{fig:model_illustration} illustrates the functional differences between the CQRNN(green) from Section \ref{CQRNN}, the MCQRNN(blue) from Section \ref{MCQRNN}, and the SCQRNN(red) proposed in the current Section. The MCQRNN needs a single quantile $\tau_i$ beeing passed in the input, together with the original data $x$. After the forward pass, this $\tau_i$ is then given to the loss function $EQ_{\tau_i}$ (Equation \ref{error_eq}), together with the one-dimensional output of the model. The gradient on $EQ_{\tau_i}$ is then used for the back propagation, denoted with a dashed arrow. The CQRNN and the SCQRNN both do not need any additional input apart from the original $x$. While both also use a $T$-dimensional output, the CQRNN uses the latter directly for the computation of the loss function $ECQ_{\tau}$ (Equation \ref{error3_eq}). The SCQRNN meanwhile sorts the output, before passing it to the $ECQ_{\tau}$. This leads to the error being propagated back to the model through the sorting operation, which is again denoted by the dashed arrow.
        While the MCQRNN uses its modified linear layers, to ensure the monotony in the $\tau$ input, the functionality of the CQRNN and the SCQRNN is not limited to an MLP-infrastructure. As long, as the underlying model ensures a $T$-dimensional output, both methods can be used to train it.
        A trained CQRNN model can also be sorted during evaluation. We showcased this post hoc approach in Section \ref{setup1}, where we called it CQRNNse.

    \subsection{Theoretical Complexity Analysis}\label{theoretical_complexity}
        In this Subsection, we will show, that the forward pass of the SCQRNN has a significant better computation complexity, than the MCQRNN. Therefore, it is in general more sustainable than the MCQRNN, both during training and evaluation. 

        We will compare the forward pass of the SCQRNN and MCQRNN for a single sample. Assume both models have $K$ hidden layers with a maximum of $L$ neurons per layer and that there are $T$ quantiles to predict.
        
        \textbf{SCQRNN:}
            First we look at the forward pass for a single layer in the SQRNN and get
            \begin{equation}\label{simple_layer}
                z_{\textbf{out}}=f(W z_{\textbf{in}}) \in \mathcal{O}(L^2)
            \end{equation}
            since the $W z_{\textbf{in}}$ is at most the multiplication between a $L\times L$ dimensional matrix and a $L$ dimensional vector and $f$ is an activation function, which usually has a linear runtime.
            By running through $K$ hidden layers, we get the complexity of $\mathcal{O}(K L^2)$ plus the complexity of the final linear layer and the sorting operation
            \begin{equation}
                \hat{y}^\tau=\mathcal{S}(f(W^{(K)}~z^{(K)})) \in \mathcal{O}(LT+T\log{(T)})
            \end{equation}
            since $W^{(K)}$ is at most a $T\times L$ matrix and sorting a $T$-dimensional vector with the algorithm of Blondel~et~al.~\cite{blondel2020fast} has the complexity of $\mathcal{O}(T\log{(T)})$. So the final complexity for the SCQRNN is $\mathcal{O}(KL^2 + LT + T\log(T))$.
      
        \textbf{MCQRNN:}
            The forward pass through a single layer of the MCQRNN looks a little different:
            \begin{equation}
               z_{\textbf{out}}=f(\exp{(W)} z_{\textbf{in}}) \in \mathcal{O}(L^2) 
            \end{equation}
            The exponential function runs in linear time and $W$ has at most $L^2$ entries. Therefore the computation of $\exp{(W)}$ stays in $\mathcal{O}(L^2)$ and the rest is equivalent to (\ref{simple_layer}). As before, by running through $K$ hidden layers, we get the complexity of $\mathcal{O}(K L^2)$, but without additional sorting in the last layer. Finally we have to consider, that a single passthrough of a sample isn't enough for the MCQRNN to train or evaluate it on all $T$ quantiles. In fact, a single original sample has to pass the MCQRNN exactly $T$ times. Therefore, the final complexity of the MCQRNN is $\mathcal{O}(TKL^2)$

        \textbf{Comparison:}
            To compare the complexity of The SCQRNN and the MCQRNN, let us assume, that the number of quantiles $T$ and the maximum layer size $L$ are proportional to eachother ($T\in\mathcal{O}(L)$). This assumption is reasonable, since in practice their sizes should not differ in a large magnitude. Also assume, that K is constant for simplicity reasons. Then the complexity of the SCQRNN collapses to $\mathcal{O}(KL^2+LL+L\log(L))=\mathcal{O}(L^2)$ and the complexity of the MCQRNN collapses to $\mathcal{O}(LKL^2)=\mathcal{O}(L^3)$.
            So while the MCQRNN has a cubic runtime, the SCQRNN only has a quadratic one, which makes it significantly faster. 
            
            
    \subsection{Datasets}\label{data}
\begin{figure}[t!]
                \centering
                \begin{subfigure}{.49\textwidth}
                    \centering
                    \includegraphics[width=.9\linewidth]{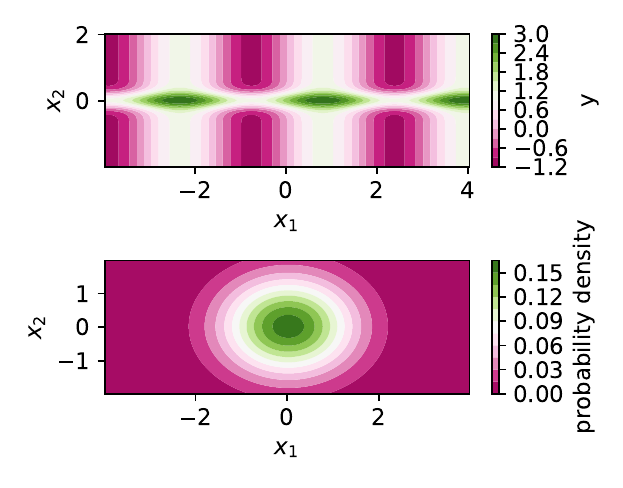}
                    \caption{Example 0}
                    \label{fig:sfig0}
                \end{subfigure}%
                \begin{subfigure}{.49\textwidth}
                    \centering
                    \includegraphics[width=.9\linewidth]{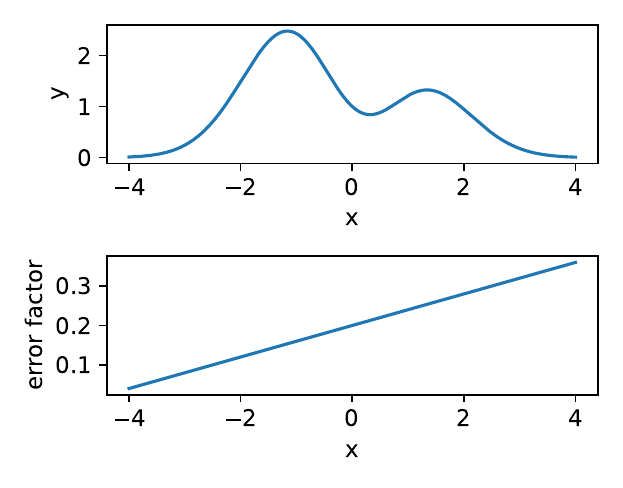}
                    \caption{Example 1}
                    \label{fig:sfig1}
                \end{subfigure}
                \begin{subfigure}{.49\textwidth}
                    \centering
                    \includegraphics[width=.9\linewidth]{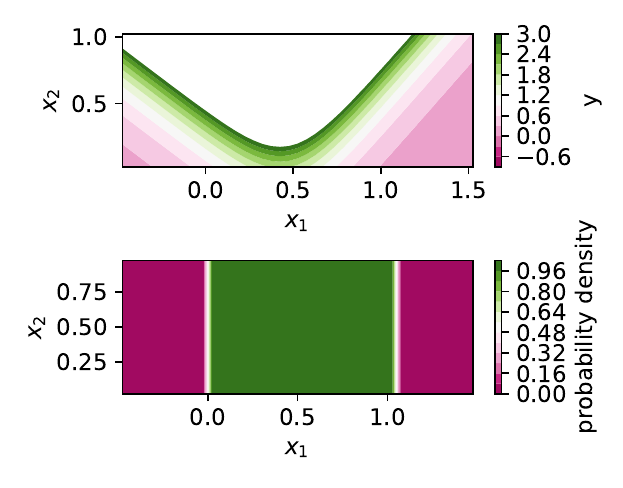}
                    \caption{Example 2}
                    \label{fig:sfig2}
                \end{subfigure}%
                \begin{subfigure}{.4\textwidth} 
                    \centering
                    \rotatebox{-90}{\includegraphics[width=.7\linewidth]{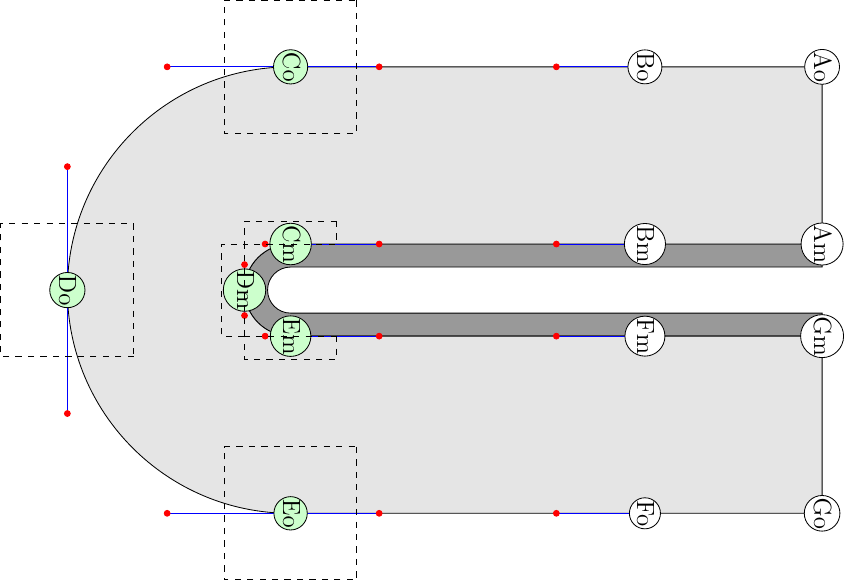}}
                    \caption{U-bend dataset. Figure taken and adapted from~\cite{decke2022ndnet}} 
                    \label{fig:sfig3} 
                \end{subfigure}%
                \caption{Visualization of the datasets used in this article}
                \label{fig:datasets}
            \end{figure}

            We use a total of ten datasets for the evaluation of our experiments. For our Experiment 1 which is detailed later in Section~\ref{setup1} we utilize three base example functions, which then get augmented with three differently distributed errors. The functions are depicted in Equations~\ref{ex0} to \ref{ex2} and illustrated in Figures~\ref{fig:sfig0} to \ref{fig:sfig2}. These base functions were originally introduced by Xu et al.~\cite{xu2017composite} and have also been utilized by Cannon~\cite{Cannon2017non}.
            \begin{equation}\label{ex0}\tag{example 0}
                y=\sin(2x_1)+2\exp(-16x_2^2)+0.5\varepsilon
            \end{equation}
            with $x_1\sim N(0,1)$ and $x_2\sim N(0,1)$;
            \begin{equation}\label{ex1}\tag{example 1}
                y=(1-x-2x^2)\exp(-0.5x^2)+\frac{1+0.2x}{5}\varepsilon
            \end{equation}
            with $x\sim U(-4,4)$ ;
            \begin{equation}\label{ex2}\tag{example 2}
                y=\frac{40\exp\{[(x_1-0.5)^2+(x_2-0.5)^2]\}}{\exp\{8[x_1-0.2)^2+(x_2-0.7)^2]\}+\exp\{8[(x_1-0.7)^2+(x_2-0.7)^2]\}}+\varepsilon
            \end{equation}
            with $x_1\sim U(0,1)$ and $x_2\sim U(0,1)$.

            In Figure \ref{fig:datasets} these Functions are visualized. For example 0 and example 2 there is a heat map of the functions in the upper plot and a heat map of the distribution of $x_1$ and $x_2$ in the lower plot. The function of example 1 has only a one dimensional input and is therefore depicted with its graph in the upper plot. The lower plot shows the scaling factor of the epsilon in terms of $x$. Note, that this makes the resulting data heteroscedastic for example 1.

            By incorporating the error term $\varepsilon$, we augment the three base functions with three distinct error functions, resulting in a total of nine datasets derived from our three base functions. The random errors $\varepsilon$ are generated from three distributions: the normal distribution with a variance of 0.25, denoted as $\varepsilon~\sim~N(0,0.25)$; the Student's $t$ distribution with three degrees of freedom, denoted as $\varepsilon~\sim~t(3)$; and the chi-squared distribution with three degrees of freedom, denoted as $\varepsilon~\sim~\chi^2(3)$. The use of the selected example functions and their associated error terms is pivotal because it enables the calculation of true quantiles. This capability is crucial enabling a reliable evaluation of our models. This is explained in more detail in the evaluation paragraph in Section~\ref{setup1}. For each combination of base functions and error distributions, we generate 600 samples, which are then evenly divided into training, testing, and validation datasets, each containing 200 samples.

            The U-bend dataset introduced by Decke et al.~\cite{decke2023dataset} is a more complicated and real-world dataset from the field of design optimization. The design of each U-bend sample is described by 28 parameters serving as the models input. This parameterized U-bend is depicted in Figure~\ref{fig:sfig3}. The points depicted in green, which can vary within the dashed boxes, describe the boundary of a design, while the red dots illustrate the Bezier parameters, indicating how the boundary points are connected. This dataset was selected to demonstrate that the SCQRNN is not limited to predicting simple mathematical functions but is also capable of addressing complex real-world problems.



    \subsection{Experiment 1}\label{setup1}
        To evaluate the performance of the SCQRNN and compare it to existing models, we use a Monte Carlo simulation based on the setup introduced by Xu et al.~\cite{xu2017composite}, which is also used by Cannon~\cite{Cannon2017non}.
        
        \textbf{Setup:}
            We consider four models for comparison: The SCQRNN (as described in Section~\ref{SCQRNN}), the MCQRNN (Section~\ref{MCQRNN}), the CQRNN (Section~\ref{CQRNN}), and the CQRNNse. Notably, the CQRNNse mirrors the CQRNN in structure but incorporates post hoc sorting during evaluation, meaning both models utilize the same underlying trained model.

            The architecture for all considered models consists of two hidden layers. Specifically, for function \eqref{ex0}, each layer comprises four neurons, while for functions \eqref{ex1} and \eqref{ex2}, the layers are configured with five neurons each~\cite{xu2017composite,Cannon2017non}. The models' objective is to predict a series of quantiles $\tau=(\tau_1,...,\tau_{19})$, with $\tau_i = 0.05i $.  Optimization for the SCQRNN, CQRNN, and CQRNNse employs the PyTorch Adam algorithm, featuring a learning rate of 0.01 and a weight decay of 0.05. Training and validation proceed in batches of 16, incorporating an early stopping mechanism triggered by validation error. The MCQRNN's optimization strategy utilizes the Adam algorithm as implemented in the qrnn CRAN~\cite{Cannon2024qrnn} package.

\textbf{Evaluation:}
Simulations are conducted 100 times, with each of the four models being fitted on each of the nine training sets and evaluated on their respective test sets. 

\begin{enumerate}
    \item \textbf{Root Mean Square Error (RMSE)}: To compute the RMSE of our predictions $\hat{y}^\tau$, we initially identify the true quantiles of our random errors using the quantile functions of their distributions. These true quantiles replace the $\varepsilon$ in the example functions to establish our ideal estimator $\accentset{\ast}{y}^\tau$. The RMSE between $\accentset{\ast}{y}^\tau$ and $\hat{y}^\tau$ provides a precise measure of our predictions' proximity to the actual dataset quantiles. This RMSE calculation, tailored to our predefined functions and error distributions, is not directly transferrable to real-world problems, as such specific information is typically rarely to never known. This Approach differs from  the method, that is used by Xu et al.~\cite{xu2017composite} and Cannon~\cite{Cannon2017non}. The RMSE, they presented for the CQRNN and MCQRNN is obtained by evaluating the conditional mean of the predicted quantiles and calculating the RMSE between this mean and the target value.
    \item \textbf{Overall Reliability}: Introduced by Gensler~\cite{gensler2019wind}, this metric assesses the observed frequency of targets in $y$ that fall below the predicted quantile function $\hat{y}^{\tau_i}$. The observed frequency $v^\tau_i$ is calculated as follows:
    \begin{equation}
        v^{\tau_i} = \frac{1}{N} \sum\limits_{n=1}^N H(\hat{y}^{\tau_i}_n - y_n)
    \end{equation}
    where $H$ represents the Heaviside step function. For an accurate estimator, the observed frequency $v^\tau_i$ should closely align with $\tau_i$. The overall reliability for a multi-quantile estimator is given by:
    \begin{equation}
        \bar{v}^\tau = \frac{1}{T} \sum\limits_{i=1}^T |v^{\tau_i} -\tau_i|
    \end{equation}
    Unlike RMSE, overall reliability is calculable with purely observational data, making it more suitable for evaluating real-world application performance. However, as noted by Gensler~\cite{gensler2019wind}, reliability does not measure regression performance but rather the statistical soundness of a predicted distribution.
\end{enumerate}
      
    \subsection{Experiment 2}\label{setup2}

        In our second experiment, we'll explore if sorting reduces epochs needed for convergence during training. Proposition~\ref{prop:pinball}  in Section~\ref{diff_sort} shows, that sorting generally decreases the pinball loss of an estimator  and even strictly decreases it in the quantile crossing cases. This mechanism is expected to provide the SCQRNN with a competitive advantage over the traditional CQRNN.

        \textbf{Setup:}
                    In this experiment, we assess the validation losses of the SCQRNN and the CQRNN using the U-bend dataset, as illustrated in Figure~\ref{fig:sfig3}. Each model has three hidden layers containing 600, 300, and 150 neurons, respectively, and aims to predict a sequence of quantiles $\tau=(\tau_1,...,\tau_{19})$, where each $\tau_i$ equals $0.05i$. Both models are optimized using Adam with a learning rate of 0.0001 and a weight decay of 0.005. They are trained and validated with a batch size of 16. The training stops, when the validation loss falls below a threshold of 0.05.

        \textbf{Evaluation:}
            We track the validation curves of 1000 iterative runs of the SCQRNN and the CQRNN. For each iteration, a consistent new random seed is applied to both models to ensure identical initial weights for every run. Subsequently, we assess the number of epochs required by each model to meet the threshold, noting the faster model. Finally, we analyze and compare both the average and median number of epochs necessary to reach the specified threshold and the associated standard deviations.

\section{Results and Discussion}\label{results}
    In this section, we present the results of the two experiments, which are described in Section \ref{setup1} and \ref{setup2}.
    \subsection{Experiment 1}
\begin{figure}[b!]         
            \centering
            \vspace{-0.4cm}
            \includegraphics[width=.6\linewidth]{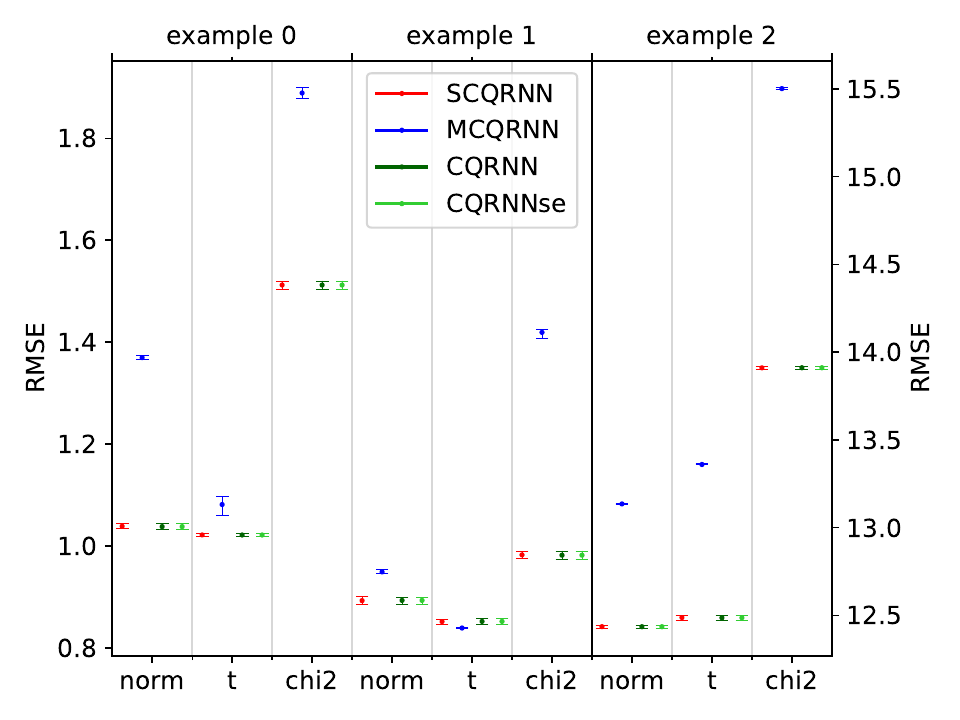}
            \vspace{-0.4cm}
            \caption{Test RMSE for four models assessed across three example functions, each augmented with three distinct error functions. Examples 0 and 1 are mapped to the left axis, whereas example 2 is scaled to the right axis}
            \label{fig:RMSE}
        \end{figure}

\begin{figure}[b!]
            \centering
            \vspace{-0.4cm}
            \includegraphics[width=.6\linewidth]{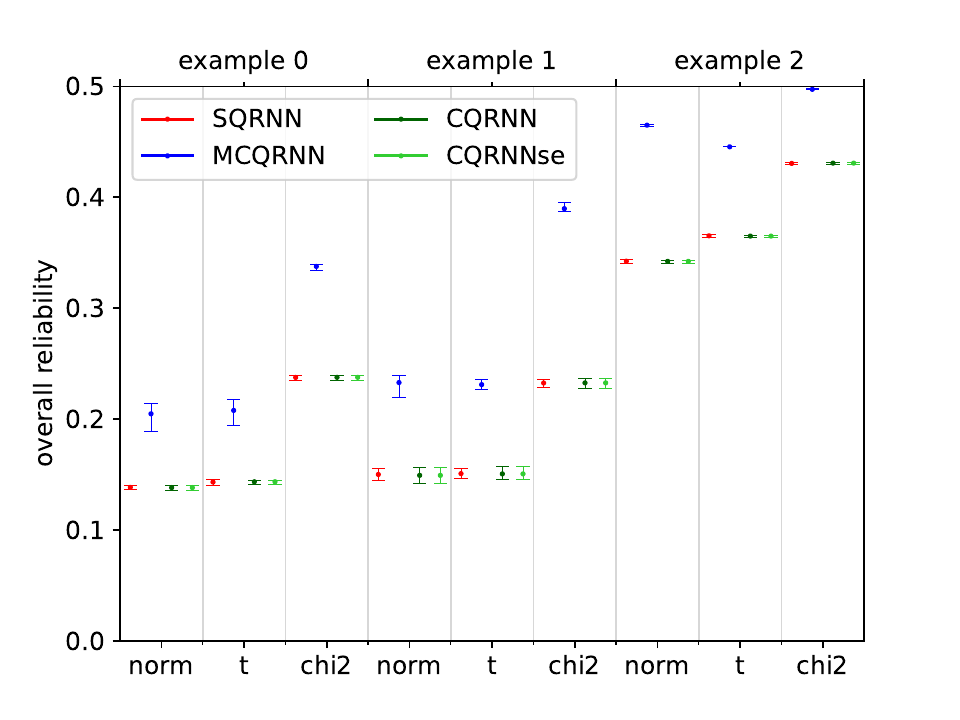}
            \vspace{-0.4cm}
            \caption{Test Overall Reliability for four models assessed across three example functions, each modified with three distinct error functions.}
            \label{fig:overall_rel}
        \end{figure}

         The observations made regarding the first experiment are visualized in the Figures \ref{fig:RMSE} and \ref{fig:overall_rel}. Figure \ref{fig:RMSE} shows the RMSE performance of the different models described in Section~\ref{setup1} on each of the 9 test datasets. These datasets consist of the three example functions with three different $\varepsilon$ values. The median RMSE, determined from 100 Monte-Carlo Iterations is depicted by the center dot, while the antennas indicate the 0.05 and the 0.95 quantile. Due to the large differences in the absolut values of RMSE, the examples 0 and 1 are aligned with the left RMSE axis and example 2 is adjusted to the right axis.
        
        The MCQRNN performs worse in RMSE compared to the SCQRNN and the baseline models, except in example 1 with $t$-distributed errors, where it outperforms them. The median RMSE between the SCQRNN and the baseline models is similar, with slight variations. There's no noticeable difference between the CQRNN and the CQRNNse. The MCQRNN is implemented in R, unlike the other models in Python with PyTorch, providing more adaptability and flexibility. Our analysis uses the original, unmodified R implementation. The highest RMSE across all examples is associated with $chi^2$-distributed errors. For examples 0 and 1, RMSE for $t$-distributed errors is slightly lower than for normally distributed errors, which is reversed in example 2. Example 2 consistently has significantly higher RMSE values.

        Figure \ref{fig:overall_rel} captures the overall reliability, mirrroring the content of Figure \ref{fig:RMSE} but with the distinction, of employing a single axis for the plot.
        As for the RMSE, the MCQRNN shows the poorest performance in the overall reliability, compared to the SCQRNN and the two baseline models, a trend that persists even for example 1 with the $t$-distributed error~$\varepsilon$.
        The difference among the remaining models is minimal, with the baseline models performing similarly. In terms of error distributions $\varepsilon$, the normally distributed $\varepsilon$ generally yields the best overall reliability, with the $t$-distributed slightly underperforming in comparison.  The $\chi^2$-distributed $\varepsilon$ shows the lowest performance. Notably, across all models, example 2 exhibits the lowest overall reliability relative to the other examples.

        The similar outcomes of CQRNN and CQRNNse suggest that post hoc sorting does not impact CQRNN performance, indicating the absence of quantile crossing during this experiment. Despite facing heteroscedastic errors, both models demonstrate competent performance, as shown in example 1. They effectively handle data with $t$-distributed errors~$\varepsilon$, indicating proficiency in managing kurtosis. However, the presence of additional skewness from the $\chi^2$-distributed errors~$\varepsilon$ may partially affect performance. Notably, with a parameter $k$ equal to 3, $\chi^2$-distributed errors~$\varepsilon$ have a mean of 3 and strict positivity, distinguishing them significantly from normally- and $t$-distributed errors~$\varepsilon$ with a mean of 0. Consequently, $\chi^2$-distributed datasets are expected to yield significantly higher values than their counterparts, leading to higher RMSE.
        
        Overall the experiment demonstrates, that in terms of RMSE and overall reliability, the SCQRNN does perform equally compared to the baseline models and notably outperforms the MCQRNN model. It is also important to note, that the SCQRNN benefits of lower computation complexity during the forward pass, as detailed in Section \ref{theoretical_complexity}. Actual runtime comparisons were not conducted in this experiment due to the disparate conditions and implementations between R and Python.

    \subsection{Experiment 2}
        The principal findings of the second experiment are presented in Table~\ref{tab:Iterations}, providing a comprehensive comparison of the epochs required to achieve the predefined loss value of $0.05$, as outlined in Section~\ref{setup2}, between the SCQRNN model and the CQRNN baseline model.
        \begin{table}[b!]
            \centering
            \caption{Experiment 2 Results: Summary of the median, mean, and standard deviation for the epochs required to reach the threshold value across 1000 simulation runs, and a counter of faster convergence runs between models.}
            \begin{tabular}{|c||c|c|c|c|}
                 \hline
                  \multirow{2}{*}{\textbf{model}} & \multicolumn{3}{c|}{\textbf{epoch}} & \textbf{converged} \\
                  \cline{2-4}
                 &  \textbf{median} & \textbf{mean} & \textbf{std} & \textbf{faster} \\
                 \hline\hline
                 SCQRNN & 64 & 63.321 & 6.329 & 995/1000\\
                 CQRNN & 75 & 75.336 & 7.895 & 2/1000\\
                 \hline
            \end{tabular}
            \label{tab:Iterations}
        \end{table}

        The SCQRNN only needs 64 epochs in median to reach the loss threshold, in contrast to the CQRNN’s 75 iterations. This translates to a 14.67\% reduction in epochs needed for the SCQRNN. When examining mean values, the SCQRNN necessitates 15.95\% fewer epochs. Furthermore, the SCQRNN exhibits a 19.84\% lower standard deviation.
        Notably the SCQRNN achieved faster convergence than the CQRNN in 995 of 1000 simulation runs. 

         \begin{figure}[t!]
            \centering
            \includegraphics[width=0.6\linewidth]{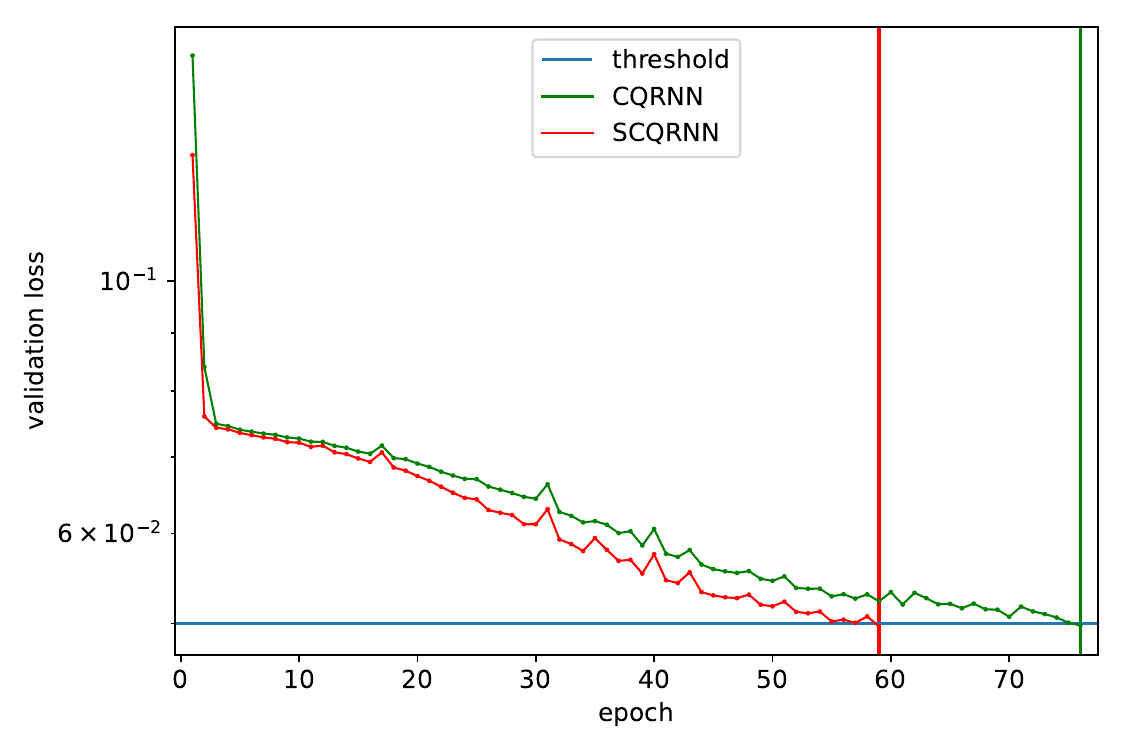}
            \vspace{-0.4cm}
            \caption{Exemplarily chosen validation curves of a single simulation run.}
            \label{fig:val_curve}
         \end{figure}

        Figure \ref{fig:val_curve} exemplarily shows a validation curve of the CQRNN and the SCQRNN in a single representative simulation run. The similarity in this curves is evident, with the curve of the SCQRNN consistently positioned below the curve of the CQRNN, indicating earlier threshold attainment.
        This similarity highlights the almost identical nature of the models, in combination with fixed seeds resulting in identical initial weights for each run. The differentiating characteristic resides in the sorting mechanism of the SCQRNN, implying it to be the key mechanism behind its faster convergence. Moreover, Proposition \ref{prop:pinball} suggests the possibility of quantile crossing with the CQRNN during validation.
        
        This experiment showed that the SCQRNN converged significantly faster in 99.5\% of runs, needing 15\% fewer epochs on average, confirming Proposition \ref{prop:pinball}'s theoretical anticipation with practical evidence.
        
\section{Conclusion}\label{conclusion}
    This article introduced the Sorting Composite Quantile Regression Neural Network (SCQRNN), a novel model designed to efficiently address the challenge of quantile crossing in neural network-based quantile regression, while significantly enhancing computational efficiency with the help of ad hoc sorting. Specifically, we demonstrated that the SCQRNN processes a sample in $\mathcal{O}(L^2)$ time for a maximum layer size $L$, contrasting with the MCQRNN's $\mathcal{O}(L^3)$ requirement. Following this, we noted a significant improvement in the model's convergence speed, observing that the SCQRNN requires approximately 15\% fewer epochs to converge compared to conventional models due to ad hoc sorting. This efficiency underscores the SCQRNN's dual advantage: faster convergence compared to the CQRNN, which does not inherently prevent quantile crossing, and superior computational time efficiency relative to the MCQRNN. Previously, the choice between models necessitated a compromise—opting for the MCQRNN to prevent quantile crossing at the expense of computational cost or selecting the CQRNN with the risk of quantile crossing. 

    The Python implementation leveraging PyTorch contributes to the SCQRNN's flexibility, enabling a broader range of configurations and optimizations beyond the limitations observed in traditional QRNN implementations. This adaptability is crucial for tailoring the model to diverse datasets and problem settings. 

    Furthermore, our study's analysis underscores the SCQRNN's potential for sustainability in HPC environments, a pressing concern in the era of machine learning and organic computing, where understanding the (un)certainty of model outcomes enhances the systems self-awareness, self-adaptability and resilience. By operating with lower computational complexity and faster convergence, the SCQRNN aligns with the urgent need for energy-efficient computational models that do not compromise on predictive performance.

    We tested our model on both high dimensional U-bend data and low dimensional example functions. We found no dimensionality-related limitations, as the sorting only affects the output layer, not the preceding MLP. Considering computational cost, the SCQRNN model adds to, rather than scales, the complexity of the CQRNN. The sorting we employ has loglinear time and linear memory complexity, which is generally dominated by the preceding MLP's quadratic complexity. The MLP can be replaced by any model with multidimensional output, potentially altering complexity.

    Future work will focus on integrating the SCQRNN into deep active design optimization~\cite{decke2023dado} (DADO), leveraging quantile regression's handling of asymmetric uncertainty and DADO's goal of finding improved samples. Since the SCQRNN is able to adapt to kurtosis and skewness, the predicted quantiles can be used to model heavy-tailed distributions. This makes the SCQRNN an ideal basis for exploring novel DADO query strategies that prioritize not just the predicted mean but also samples with a heavy left tail, identifiable through substantial median to lower quantile deviations. This approach intends to enhance sample identification efficiency by leveraging the SCQRNN's advancements.
\bibliographystyle{ieeetr}
\bibliography{literatur}

\end{document}